%% file: egpaper_for_review.tex
\documentclass[10pt,twocolumn,letterpaper]{article}
\pdfoutput=1

\usepackage{iccv}
\usepackage{times}
\usepackage{epsfig}
\usepackage{graphicx}
\usepackage{amsmath}
\usepackage{amssymb}
\usepackage{xspace}
\usepackage{multirow}
\graphicspath {{figures/}}

\usepackage[pagebackref=true,breaklinks=true,letterpaper=true,colorlinks,bookmarks=false]{hyperref}

\iccvfinalcopy 

\def\iccvPaperID{1866} 

\input{tex/header.tex}

\usepackage[numbers,sort]{natbib}
\ificcvfinal\fi
\begin{document}

\title{Scaling Object Detection by Transferring Classification Weights}

\author{
Jason Kuen$^{1}$~~~~Federico Perazzi$^{2}$~~~~Zhe Lin$^{2}$~~~~Jianming Zhang$^{2}$~~~~Yap-Peng Tan$^{1}$\\
$^{1}$Nanyang Technological University, Singapore~~~~~~~$^{2}$Adobe Research\\
}

\maketitle
\thispagestyle{empty}

\input{tex/abstract.tex}

\input{tex/introduction.tex}
\input{tex/relatedworks.tex}

\input{tex/approach.tex}
\input{tex/experiments.tex}
\input{tex/conclusion.tex}

{\small
\bibliographystyle{ieee_fullname}
\bibliography{egpaper_for_review}
}

\end{document}

%% file: tex/header.tex
\usepackage{xcolor}
\usepackage{pifont}
\newcommand{\sharedcls}{\ensuremath{S}\xspace}
\newcommand{\novelcls}{\ensuremath{N}\xspace}
\newcommand{\clsnetcls}{\ensuremath{C}\xspace}
\newcommand{\detnetcls}{\ensuremath{D}\xspace}
\newcommand{\clsnet}{\text{CLN}\xspace}
\newcommand{\detnet}{\text{DEN}\xspace}
\newcommand{\weights}[1]{\ensuremath{W_{#1}}\xspace}

\newcommand{\ap}{\ensuremath{\text{AP}_{50}\xspace}}
\newcommand{\ar}{\ensuremath{\text{AR}_{50}\xspace}}
\newcommand{\wtnplus}{\ensuremath{\text{WTN}^{+}\xspace}}
\newcommand{\xmark}{\ding{55}}%
\newcommand{\cmark}{\ding{51}}
\newcommand{\losscls}{\ensuremath{\ell_\text{cls}}}
\newcommand{\lossreg}{\ensuremath{\ell_\text{box}}}
\newcommand{\lossrec}{\ensuremath{\ell_\text{rec}}}

%% file: tex/abstract.tex
\begin{abstract}

Large scale object detection datasets are constantly increasing their size in terms of the number of classes and annotations count. Yet, the number of object-level categories annotated in detection datasets is an order of magnitude smaller than image-level classification labels. State-of-the art object detection models are trained in a supervised fashion and this limits the number of object classes they can detect. In this paper, we propose a novel weight transfer network (WTN) to effectively and efficiently transfer knowledge from classification network's weights to detection network's weights to allow detection of novel classes without box supervision. We first introduce input and feature normalization schemes to curb the under-fitting during training of a vanilla WTN. We then propose autoencoder-WTN (AE-WTN) which uses reconstruction loss to preserve classification network's information over all classes in the target latent space to ensure generalization to novel classes. Compared to vanilla WTN, AE-WTN obtains absolute performance gains of $6\%$ on two Open Images evaluation sets with 500 seen and 57 novel classes respectively, and $25\%$ on a Visual Genome evaluation set with 200 novel classes.

\end{abstract}

%% file: tex/introduction.tex
\section{Introduction}

\begin{figure}[t]
	\centering
  \includegraphics[trim=0 0 0 0,clip, width=\linewidth]{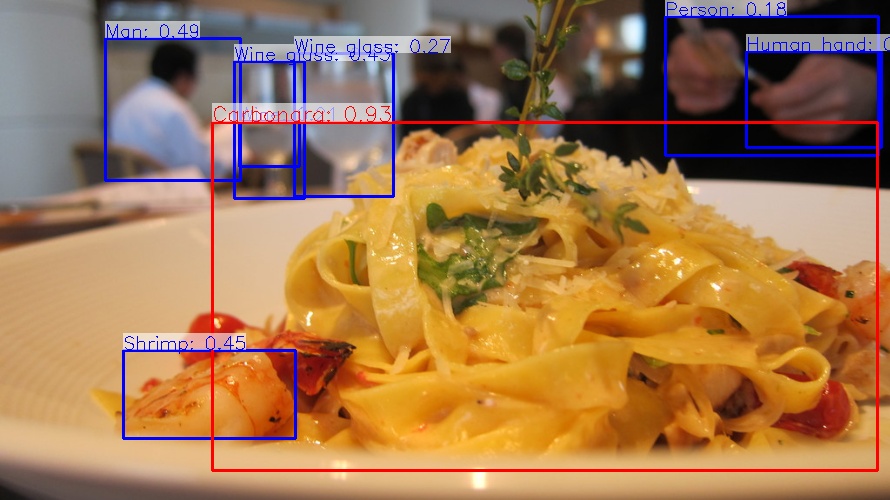}
	\caption{Our proposed detector has no access to box-level training annotations for the object class represented by the \textcolor{red}{red} box, ``\textit{Carbonara}''. It learns to detect novel object classes by transferring weight knowledge from large-scale pre-trained image classification network.}
	\label{fig:teaser}
\end{figure}

State-of-the-art object detectors \cite{ren2017faster, he2017mask} are typically trained with a large number of bounding box annotations. Large-scale datasets such as COCO \cite{lin2014microsoft}, Pascal VOC \cite{pascal} and OpenImages \cite{openimages} provide a substantial amount of bounding boxes, but the number of annotated object categories is often very limited.~The reason is that scaling the number of bounding boxes can be semi-automated, \eg \cite{openimages}, while increasing the number of classes requires significant human labor. On the other hand, image-level labels such as those available in classification datasets are much easier to collect as they do not require costly bounding box annotations.~As a consequence, several works investigated the training of object detectors in a weakly-supervised regime, using only image-level labels.~These methods leverage a variety of classes available in classification datasets or image tags found in social networks \cite{MahajanGRHPLBM18} but neglect the spatial information available in object detection datasets.

In contrast, partial supervised methods \cite{hu2018learning} employ both types of annotations. While existing methods \cite{li2016weakly,redmon2017yolo9000,tang2016large,tang2018visual} that transfer knowledge from a classification network to a detection network with partial supervision achieve higher accuracy than weakly-supervised methods \cite{li2016weakly,bilen2016weakly,bilen2015weakly}, they incur a significant computational cost during training and testing. The overhead comes either from joint training of the two networks \cite{li2016weakly,redmon2017yolo9000}, or from performing forward passes of the classification network during testing \cite{tang2016large, tang2018visual}. Furthermore, joint-training methods often require storage-intensive, large-scale classification datasets to be present while training the detection network.

To overcome these limitations, we propose a novel approach to transfer discriminative semantic knowledge from classification to detection with a non-linear weight-transfer network (WTN) \cite{hu2018learning}. Given a set of common classes annotated for both tasks, we learn a function, the weight-transfer network, that maps weights at the fully-connected layer of the classification network to those of the object detection network. Once trained, WTN is used to extend the number of categories recognized by the object detector via transferring weights of unseen classes from the classification network.
This strategy is advantageous because it only adds little computational and memory overheads to training and no burden to inference at all.

Compared to the vanilla weight-transfer network \cite{hu2018learning}, we introduce two key components to our model. First, we insert normalization layers to account for the different amplitude of the classification weights. Secondly, we replace the multilayer perceptron with an autoencoder. The latent space of the autoencoder corresponds to the classification weights of the object detector and therefore is trained with object-level supervision. The reconstruction loss between the input and output of the autoencoder is essential to retain semantic information of all the classes while the detection network's classification loss facilitates the learning of a discriminative embedding of the class weights.

Extensive experimentation on Open Images \cite{openimages} and Visual Genome \cite{krishna2017visual} datasets demonstrates that the proposed method significantly outperforms existing partially-supervised detection approaches on challenging detection tasks involving novel object classes. Moreover, due to the auxiliary regularization effect brought by the reconstruction loss of autoencoder WTN, our proposed method even recovers the performance loss of existing WTN on seen classes.

\bigskip
\noindent\textbf{Contributions}.~The contributions of this work are threefold: \emph{i)} we address the under-fitting issue of WTN by introducing input and feature normalization schemes. The resulting model \wtnplus achieves improved detection performance over the vanilla WTN; \emph{ii)} we propose our main model, autoencoder WTN, that better preserve semantic knowledge of all object classes, while learning to generate discriminative classification weights for the detection network; \emph{iii)}~we verify the effectiveness of our method with extensive evaluations using large-scale datasets with millions of images and several hundreds of object classes.

%% file: tex/relatedworks.tex
\section{Related Works}
Over the years, several convolutional network-based object detection frameworks and architectures have been proposed: R-CNN \cite{girshick2014rich}, Fast R-CNN \cite{girshick2015fast}, Faster R-CNN \cite{ren2017faster}, R-FCN \cite{dai2016r}, SSD \cite{liu2016ssd}, YOLO \cite{redmon2016you,redmon2017yolo9000}, FPN \cite{lin2017feature}. They can be roughly categorized into single-shot detectors \cite{dai2016r,liu2016ssd,redmon2016you,redmon2017yolo9000} which predict detection boxes from feature maps directly, and two-shot detectors \cite{girshick2014rich,girshick2015fast,ren2017faster} which first generate object proposals and then perform spatial extraction of feature maps based on the proposals for further predictions. These approaches have improved object detection from an algorithmic perspective and in a fully supervised setting.  In this work, we adopt Faster R-CNN \cite{ren2017faster} because its box-level \textit{classification head} learns just a single set of classification weights, resembling image-level classification (source task) networks. This allows a smoother knowledge transfer from classification to detection, compared to using single-shot detection networks which learn multiple sets of classification weights for different anchor boxes.

Object-levels annotations are time-consuming and tedious to collect, especially when the number of classes is large. With a large number of classes, it is very challenging to obtain accurate and complete annotations due to complex overlapping meanings of classes. Thus, several approaches attempt to scale up the number of object classes handled by object detectors using image-level annotations. Transferring knowledge from image classification to object detection is an active research area tackling the lack of bounding box annotations of the target datasets and/or object classes. These knowledge transfer-based methods for scaling up object detection can be divided into two categories: weakly-supervised and partially-supervised approaches.

Weakly-supervised methods typically rely only on an image-level classification dataset and leverage class agnostic box proposals or prior object knowledge to build object detectors. For example, Uijlings \etal \cite{uijlings2018revisiting} perform multiple instances learning with knowledge transfer (source dataset with bounding boxes) to produce boxes for the target training dataset. In \cite{tao2018zero}, a weakly-supervised object detector is trained on a weakly-labeled web dataset to generate pseudo ground-truths for the target detection task. \cite{singh2018r} combines region-level semantic similarity and common-sense information learned from some external knowledge bases to train the detector with just image-level labels.

More closely related to our work are weight adaptation methods \cite{hoffman2014lsda,tang2016large,tang2018visual} that fine-tune classification networks and learn detection-specific bias vectors to adapt the networks for detection. These adaptation-based methods assume the classification power of the network is well-preserved (e.g., using R-CNN \cite{girshick2014rich}) when transferred to the detection task. This restricts them from being effectively applied to recent detection methods (e.g., Faster R-CNN \cite{ren2017faster}, feature pyramid network \cite{lin2017feature}) that significantly modify the backbone network structure. Whereas, our method is not restricted by such constraints.

In general, classification weight-based knowledge transfer \cite{hu2018learning} can be applied to any recent detection frameworks \cite{liu2016ssd,ren2017faster,redmon2017yolo9000}. On the other hand, partially supervised approaches employ weak labels, i.e. image-level annotations, as well as bounding box-level annotations. For example, YOLO-9000 \cite{redmon2017yolo9000} extend the detector's class coverage by concurrently training on bounding box-level data and image-level data, such that the image-level data contribute only to classification loss. By decoupling the detection network into two branches (positive-sensitive \& semantic-focused), R-FCN-3K \cite{singh2018r} is able to scale detection up to 3000 classes despite being trained on limited bounding box annotations for several object classes. In contrast to these, we focus on large-scale object detection without having access to additional data (classification) sources during the training. A well-trained image classification network possesses sufficiently rich semantic knowledge about the large-scale dataset's categories and the information is compressed in weights of its classification layers. We argue that such weights can effectively be exploited to help build an object detector handling a large number of categories.

%% file: tex/approach.tex
\section{Weight Transfer Network}
\noindent{\textbf{Preliminaries}.}~We consider the setting of a classification network \clsnet that handles object classes \clsnetcls, and a detection network \detnet that handles object classes \detnetcls. The number of categories handled by \clsnet is much greater than the number of categories handled by \detnet, i.e. $|\clsnetcls| >> | \detnetcls|$. The goal of our approach is to expand the number of categories handled by \detnet through partial supervision, where we transfer weight knowledge from \clsnet (source task) to \detnet (target task). We make use of the final fully-connected (FC) layer weights of the \clsnet that has been pre-trained on a large scale image classification dataset. The final FC layer weights can be seen as a form of \textit{semantic embeddings} comprising rich knowledge about the object categories and the complex class relationships. Furthermore, pre-trained large-scale image classification networks are very accessible and many are shared publicly.

Classification knowledge from \clsnet is transferred to \detnet using a weight transfer network (WTN) through the object categories shared (\sharedcls) between the two tasks: $\sharedcls = \clsnetcls \cap \detnetcls$. WTN is a neural network that works as a class-generic function $T()$ used to transform per-class classification weight vectors  $\weights{\clsnetcls}=[w^1_\clsnetcls, w^2_\clsnetcls, ..., w^{|\clsnetcls|}_\clsnetcls]$ from \clsnet to \detnet's classification weights $\weights{ \detnetcls}=[w^1_\detnetcls, w^2_\detnetcls, ..., w^{|\detnetcls|}_ \detnetcls]$ as follows: $\weights{ \detnetcls} = T(\weights{\clsnetcls})$.

WTN is trained jointly with \detnet on detection dataset with classes \detnetcls. The gradients of WTN's network parameters come from \detnet's box-level classification loss \losscls. Before training WTN and \detnet, we `freeze' \weights{\clsnetcls} (taken from pre-trained \clsnet). While \sharedcls rely on WTN, for the \detnet's categories which are not part of \sharedcls (i.e., \detnetcls $\setminus$ \sharedcls), we train their weights as in conventional detection network. To obtain \detnet's classification score predictions, we simply perform matrix multiplication between \detnet's box-level visual features and WTN's predicted weights, just like how it works for conventional classification weights. Conventionally, WTN is based on a two-layer multi-layer perceptron (MLP) architecture.

Due to its class-genericness, WTN is able to carry out effective \textit{inductive learning} \cite{de1998machine}. In other words, despite that only classes \sharedcls are seen by WTN and \detnet during training, during testing WTN (and the \detnet model that incorporates WTN) can work reasonably well with classes \novelcls of \clsnet that are not shared with \detnet, i.e. $ \novelcls = \clsnetcls \setminus \sharedcls$.

\begin{figure}[t]
	\centering
	\includegraphics[width=0.9\linewidth]{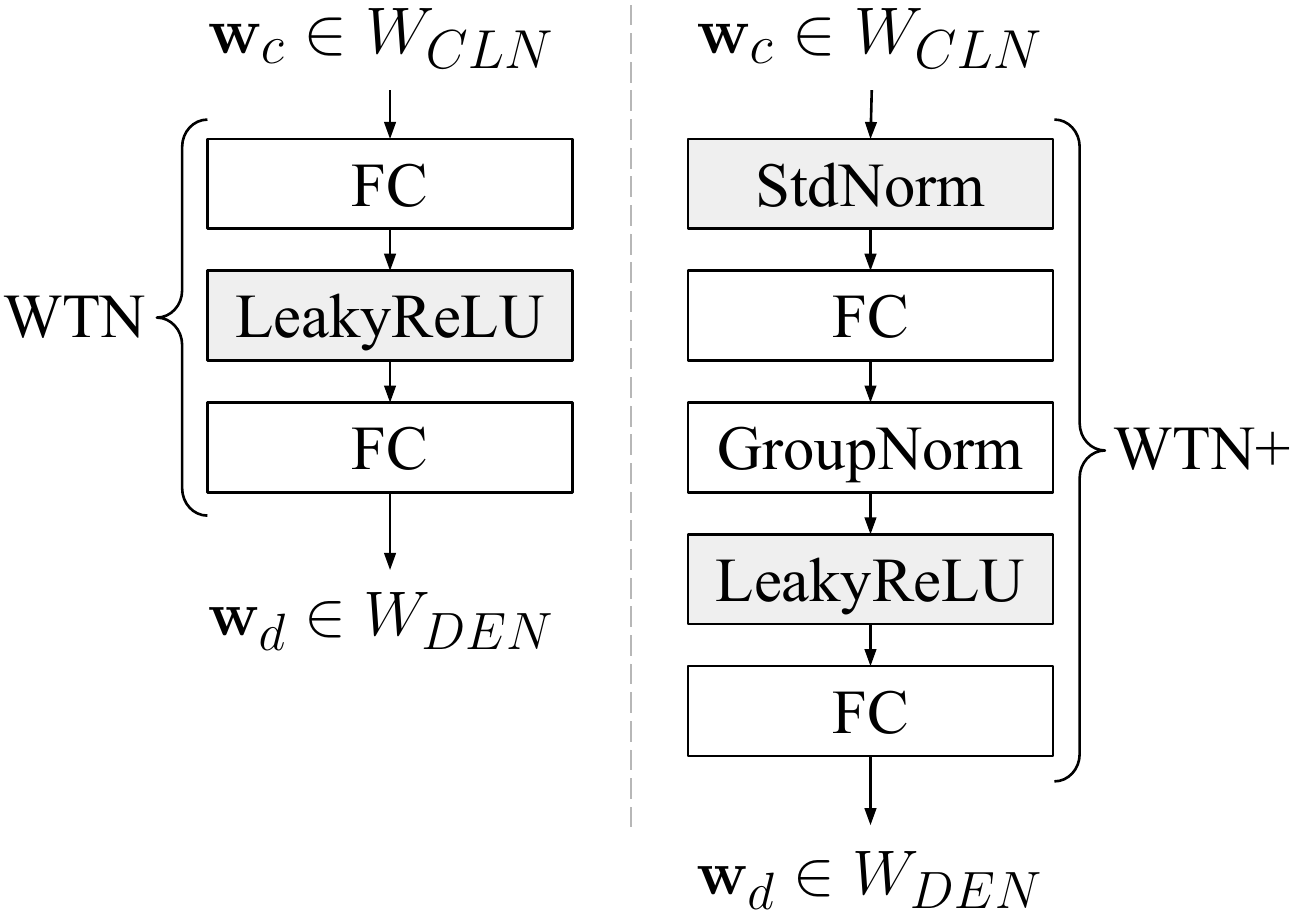}
	\caption{Comparison between network architectures of WTN  and \wtnplus. The white rectangles correspond to layers with learnable parameters.}
	\label{fig:wtnplus}
\end{figure}
\begin{figure}[t]
	\centering
	\includegraphics[width=1\linewidth]{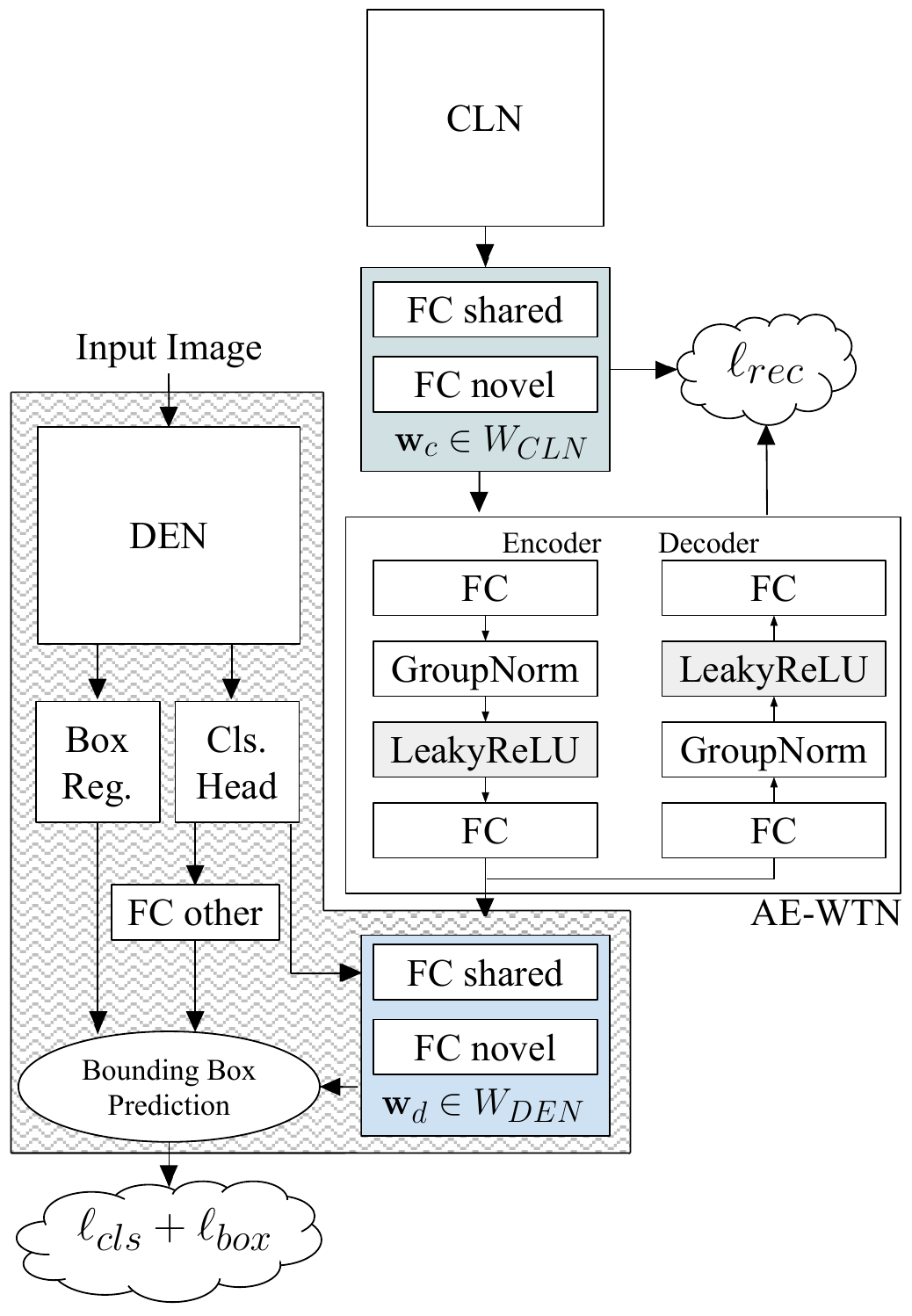}
	\caption{The \textit{train} and \textit{test} phases of object detector (DEN) with an Autoencoder-WTN (AE-WTN). \textit{Train phase}: Before training \detnet, we extract \clsnet's final FC layer's weights \weights{\clsnetcls}, and discard the earlier layers. Trained simultaneously with \detnet, AE-WTN learns to transform weights from \clsnet to \detnet through the shared classes \sharedcls. The ``other'' detection classes (i.e., $\detnetcls\setminus\sharedcls$) are trained normally as conventional classification weights. Only ``other" and \sharedcls contribute to the detection loss \losscls. AE-WTN uses a reconstruction loss \lossrec \xspace to reconstruct the weights for both \sharedcls and \novelcls, from its encoder's outputs.~\textit{Test phase}~(dashed polygon): \clsnet's weights of both the novel classes \novelcls and shared classes \sharedcls can be adapted offline for use in \detnet through AE-WTN. With that, \detnet is able to detect novel classes \novelcls in addition to \sharedcls and ``other" classes.}

	\label{fig:aewtn}
\end{figure}
\paragraph{\textbf{Normalizations}.}~Large-scale classification datasets have an unbalance class distribution, which has strong implications in how the classification weights of \clsnet are trained. E.g., in one large-scale \clsnet, we discover that the `highest-norm' class has a weight vector \textit{norm} that is 28 times that of the `lowest-norm' class. Besides, a \textit{class-generic non-linear} WTN naturally cannot adapt and learn as well as (conventional) \textit{class-specific linear} classification weights, for loss minimization. These pose challenges to the training and optimization of WTN. Empirically, we found that training  a detection network (\detnet) with existing WTN methods deteriorates the performance on \detnetcls classes, compared to a conventional \detnet trained on the same labels but without WTN.

Thus, drawing from the recent findings in activations normalization techniques \cite{ioffe2015batch, wu2018group}, we introduce a new variant of WTN,~\wtnplus~that improves performance on \detnetcls classes and it is easier to optimize. The model architectural differences between WTN and \wtnplus are illustrated in Fig.~\ref{fig:wtnplus}. Standard normalization is applied to the input weights \weights{\clsnetcls} to enable different input channels to contribute comparably to the prediction of \weights{ \detnetcls}, in order to curb the overdominance/underdominance of certain categories. Let $v_j$ denote the weights of $j$-th feature/channel of \weights{\clsnetcls}, we normalize $v_j$ by: $ \frac{v_j - \mu(v_j)}{\sigma(v_j)} $, where $\mu(\cdot)$ and $\sigma(\cdot)$ are the \textit{mean} and \textit{standard deviation} functions respectively. Group Normalization \cite{wu2018group} layer, known for its strong optimization benefits, is added to normalize intermediate features to encourage good gradient flows for easier network optimization. These small but crucial modifications are the key to training highly effective WTN.\label{sec:wtnplus}

\section{Autoencoder Weight Transfer Network}
During training, only the shared classes \sharedcls contribute to the gradients and losses of WTN. The novel object classes \novelcls are unknown to and unconsidered by WTN. The lack of knowledge of the entire class population of \clsnetcls limits WTN's capability to effectively model the \textit{good} classification space originally attained by the pre-trained \clsnet for handling a large number of categories. We hypothesize that by letting WTN have a narrow view of the class population, its modeling capability (relating to  \novelcls specifically) is severely underexploited and this compromises the performance of WTN on classes \novelcls.

To this end, we introduce Autoencoder-WTN (AE-WTN) -- a novel WTN variant that attempts to preserve knowledge on all of classes \clsnetcls contained in pre-trained \weights{\clsnetcls}, while learning a discriminative WTN function to achieve good detection performance. AE-WTN is an autoencoder with both encoder and decoder networks. AE-WTN is built on top of \wtnplus. The encoder network shares the same architecture as \wtnplus's, while the decoder network (with separate network layers/parameters) is the mirrored version of the encoder. Following existing WTN, the encoder network works as a function $T()$ to predict \weights{ \detnetcls} given \weights{\clsnetcls} as input. During training, gradients are propagated from \detnet's loss to the encoder network. The network architecture of AE-WTN and how it interacts with \clsnet and \detnet are illustrated in Fig.~\ref{fig:aewtn}.

AE-WTN is trained with an additional autoencoder-based training loss -- \textit{reconstruction loss} \cite{hinton1994autoencoders,gu2019scene} that forces the decoder network to predict (or reconstruct) the original inputs, from the output activations of the encoder network. Let $T()$ denote the encoder network and $G()$ denote the decoder network, the reconstruction is predicted as follows: $\hat{\textbf{w}}_\clsnetcls = G(T(\textbf{w}_\clsnetcls)) ;\; \forall \textbf{w}_\clsnetcls \in \weights{\clsnetcls}$. Here, we adopt smooth L1 loss \cite{girshick2015fast} as the reconstruction loss to minimize the difference between the predicted reconstructions and the original inputs (\weights{\clsnetcls}):
\begin{equation}
    \lossrec =
        \begin{cases}
        0.5 (\hat{\textbf{w}}_\clsnetcls - \textbf{w}_\clsnetcls)^2, & \text{if } |\hat{\textbf{w}}_\clsnetcls - \textbf{w}_\clsnetcls| < 1 \\
        |\hat{\textbf{w}}_\clsnetcls - \textbf{w}_\clsnetcls| - 0.5, & \text{otherwise }
        \end{cases}
\end{equation}
\noindent Note that we apply reconstruction loss to all \clsnet classes \clsnetcls (i.e., $\sharedcls \cap \novelcls$), rather than just shared classes \sharedcls. On the other hand, the detection loss (box-level classification) only cares about classes \sharedcls and ``other" detection classes. With such formulation, we perform multi-task training based on the following mixture of training losses (excluding Region Proposal Network's \cite{ren2017faster}: $\losscls + \lossreg + \alpha\lossrec$, where \lossreg \phantom{1}is box regression loss and $\alpha$ is the loss scaling hyperparameter.

Reconstruction loss penalizes intermediate network activations which do poorly to reconstruct the original weights \weights{\clsnet}. Since AE-WTN's output \weights{ \detnetcls} (weights for \detnet) is a form of intermediate network activations, they are affected by the reconstruction loss and are expected to retain original class information greatly for reconstruction purpose. In contrast, existing WTN (or even \wtnplus) is solely driven by \detnet's classification loss (which may not be optimal for model generalization) and is not compelled to retain more of potentially useful class information. Reconstruction-based information preservation has been shown to help neural networks achieve better local optima \cite{le2018supervised,zhang2016augmenting} in supervised learning. By complementing \clsnet's classification loss with a reconstruction loss, AE-WTN is able to learn a non-linear mapping that achieves a good balance between class/class discriminability and class information retainment. We find that this has a regularization effect on AE-WTN and it helps improve generalization performance on the fully-annotated object categories ($\detnetcls \cap \sharedcls$) seen during training. This observation is aligned with the findings of \cite{le2018supervised,zhang2016augmenting} that supervised learning can be improved with autoencoders. While we apply reconstruction loss to all classes including \novelcls (which do not have supervised annotations), \cite{le2018supervised,zhang2016augmenting} apply the loss to only input examples with supervised annotations. Our work also resembles semi-supervised learning where reconstruction loss (autoencoder) \cite{zhao2016stacked,rasmus2015semi} is used as an auxiliary loss to exploit unlabeled data (in this work, class \novelcls are unlabeled) to improve model performance and generalization.

During the training of existing WTN, \weights{ \clsnetcls,\novelcls} the weights of novel classes \novelcls, contained in \weights{ \clsnetcls}, is not utilized. And, classes \novelcls do not contribute to the training. Deep neural networks are generally known to eliminate task-irrelevant information of the inputs through training \cite{tishby2015deep, shwartz2017opening}. Thus, it is likely that WTN learns to ``dismiss" some class information about classes \novelcls that is unimportant to classes \sharedcls but is useful for the detection of classes \novelcls. The reconstruction loss of AE-WTN addresses such a shortcoming of existing WTN by explicitly involving the novel object classes \novelcls. The rich class information in \weights{\clsnetcls,\novelcls} (which is potentially beneficial to AE-WTN's test-performance on classes \novelcls) is preserved in the intermediate network activations of AE-WTN.

%% file: tex/experiments.tex
\section{Experiments}
\subsection{Implementation Details}
\noindent\textbf{Training and evaluation sets for seen classes \detnetcls}. We use the official training and validation dataset (referred to as \textbf{OI-500}) \cite{openimages} from Open Images V4 Challenge which contains 500 object classes for training and evaluating \detnet on classes \detnetcls. The object classes in Open Images dataset are hierarchically organized and many classes are not mutually exclusive. Open Images' official evaluation metric \cite{openimages}, a custom version of ``Average Precision (AP) @ 0.5 IoU threshold" or \ap \hspace*{1pt} is used for evaluation on the validation set provided. We use the same Open Images training set to train baseline Faster RCNN and our WTN-based models for fair comparisons on novel classes \novelcls.\\

\noindent\textbf{Evaluation set for novel classes}. \novelcls To evaluate \detnet's performance on novel classes \novelcls, we employ two evaluation datasets. The first evaluation set  (\textbf{OI-57}) is a subset of Open Images V4 \textit{complete}/non-challenge dataset containing 57 \textit{novel} object classes and 31,061 images. The second evaluation set (\textbf{VG-200}) is set as a subset of Visual Genome \cite{krishna2017visual} dataset containing 24,690 images spanning 200 high-frequency object classes which are \textit{novel} to \detnet. We adopt the same $\text{AP}_{50}$ metric for \textbf{OI-57}. Since many object instances in Visual Genome dataset are not annotated at all, we follow the practice of \cite{bansal2018zero} by using \textit{Average Recall}/\ar@100 detections per image to gauge the detection performance of \detnet on this evaluation set.\\

\noindent\textbf{Classification Network (\clsnet)} (\textit{source}). Prior to training WTN and \detnet, a pre-trained large-scale \clsnet model has to be acquired. We use a publicly available ResNet-101 pre-trained on Open Images v2 \cite{openimages} with 5000 object classes. It is trained with multi-label (sigmoid) classification loss given the multi-label nature of the dataset. Training resolution is $299\times299$. The model is trained asynchronously with 50 GPU workers and batch size 32 for 620K training steps. Incoming features to the final classification layer is $2048$-dimensional.\\

\noindent\textbf{Detection Network (\detnet)} (\textit{target}). The \detnet architecture in this paper is a Faster R-CNN \cite{ren2017faster} with a backbone integrating ResNet-50 \cite{he2016deep} and Feature Pyramid Network (FPN) \cite{lin2017feature}. ResNet-50 backbone is pre-trained on ImageNet-1k \cite{russakovsky2015imagenet} dataset, and its BN parameters are frozen during training of \detnet. The box-level head (for box classification and regression) is a 2-layer multi-layer perceptron (MLP) with a 2048-dimensional feature and output channels. \detnet is trained with mini batches of 8 images (2 images/GPU) for a total of 180K iterations. We optimize the network using SGD with momentum of 0.9 and initial learning rate of $2\times10^{-2}$. The network is regularized with weight decay of $1\times10^{-4}$. We stick closely to the
original training loss functions of Faster R-CNN except
for the classification loss which we replace with sigmoid
binary cross-entropy, taking Open Images’ class hierarchy
and multilabel nature into account. The training class labels
are expanded \cite{akiba2018pfdet} based on the hierarchy tree \cite{openimages} given.\\

\noindent\textbf{Weight Transfer Network (WTN)}. By default, WTN variants have input/feature/output channels of 2048. For Group Normalization (GN) layer in \wtnplus and AE-WTN, we follow the same ``number of groups"/$\#$groups hyperparameter, which is set to 32 as found to be a good choice by \cite{wu2018group}. WTN networks are trained from scratch simultaneously with \detnet using AdamW \cite{loshchilov2019decoupled} using default hyperparameters and weight decay of $1\times10^{-4}$. For AE-WTN, $\alpha$ is set to 20 throughout the experiments.

\begin{table}[t!]
    \centering
   \resizebox{\linewidth}{!}{
    \begin{tabular}{l||c|c|c}
    ~                    & \textbf{OI-500} & \textbf{OI-57} & \textbf{VG-200}\\
    & (Seen) & (Novel) & (Novel)\\\hline
    \textbf{Method}               & \ap                    & \ap                   & \ar                  \\\hline
    Faster R-CNN \cite{ren2017faster}  & 59.55                          &   -                            & - \\
    Faster R-CNN (NN)  &  -                           & 28.09                              & 49.39\\
    LSDA \cite{hoffman2014lsda}                & 59.44                               & 25.89                              & 51.14\\
    LSDA (Visual Transfer) \cite{tang2016large}               & 59.44                               & 26.43                              & 53.03\\
    ZSD \cite{bansal2018zero} with \clsnet weights                 & 47.37                               & 34.63                             & 38.04\\
    ZSD \cite{bansal2018zero} with fastText \cite{joulin2017bag}                  & 58.39                               & 29.51                            & 35.09\\
    WTN \cite{hu2018learning} & 52.87 & 34.94 & 41.91\\\hline
    \wtnplus         &                           &                        & \\
	$\blacktriangleright$ default model         & 58.82                           & 39.28                          & 65.60\\
    $\rhd$ $5\times$ weight decay         & 58.46                           & 40.79                          & 65.87\\
    $\rhd$ activity regularizer \cite{merity2017revisiting}          & 55.86                          & 33.47                          & 36.26\\
    $\rhd$ Dropout \cite{srivastava2014dropout}         & 57.14                           & 40.09                          & 65.52\\
    $\rhd$ reduced capacity & 58.80                           & 37.81                          & 63.16\\\hline
    AE-WTN               & \textbf{59.59}                           & \textbf{41.07}                          & \textbf{66.75}
    \end{tabular}
    }
    \label{table:comparisonrelated}
    \caption{Comparison with weight transfer-related methods on evaluation datasets -- OI-500 (seen classes), OI-57 (novel classes), and VG-200 (novel classes).}
\end{table}

\subsection{Comparison with related methods}
To validate the effectiveness of our proposed AE-WTN model, we experimentally compare it with existing weight transfer-related methods described in the following. Note that all these methods use the same Faster R-CNN detection framework and a ResNet-50 backbone.

\noindent\textbf{$\bullet$Faster R-CNN}: Vanilla Faster R-CNN \cite{ren2017faster} performs fully supervised learning on seen classes. In contrast to WTN, vanilla Faster R-CNN learns conventional classification weights which are both linear and class-specific. To detect novel classes, we employ the nearest-neighbor approach \textit{(NN)}, taking the detections of nearest seen classes.

\noindent\textbf{$\bullet$LSDA} \cite{hoffman2014lsda}: LSDA adapts \clsnet's weights for detection task by learning additive class-specific biases. To make predictions for a novel class during test-time, the biases of nearest classes are averaged and added to \clsnet's weight vector. The visual similarity transfer variant \cite{tang2016large} is also included.

\noindent\textbf{$\bullet$ZSD} \cite{bansal2018zero}: ZSD performs zero-shot detection through pre-trained word embeddings. In a joint visual-word embedding setting, the detector learns to output visual embeddings in the words' embedding space. Here, two kinds of embeddings are considered -- \clsnet's weights and fastText \cite{joulin2017bag}.

\noindent\textbf{$\bullet$WTN} \cite{hu2018learning}: This corresponds to the standard (existing) WTN model that makes use of neither normalization techniques nor reconstruction loss.

\noindent\textbf{$\bullet$\wtnplus variants}: Since the reconstruction loss of AE-WTN can be seen as a regularizer, we compare it with several \wtnplus variants regularized with increased weight decay ($5\times$) \cite{krogh1992simple}, activity regularizer (0.01) \cite{merity2017revisiting}, Dropout (0.3) \cite{srivastava2014dropout} on intermediate activations, and reduced network capacity (halving the number of channels in hidden layer).

The results are given in Table \ref{table:comparisonrelated}. We use ResNet-50 as the backbone for the vanilla Faster RCNN detector, and its \ap \hspace{1pt} on OI-500 is $59.55\%$ which is mildly worse than the $60.0\%$ achieved by the state-of-the-art SE-ResNeXt-101 detector \cite{akiba2018pfdet}. Overall, WTN methods outperform the non-WTN methods by large margins on the novel classes (OI-57 and VG-200), due to the powerful weight transfer function learned by WTN that can generalize to many classes. Among the WTN methods, AE-WTN that incorporates all the proposed improvements achieves the best results.

WTN and \wtnplus suffer from the weakened performance on OI-500 (seen classes \detnetcls) compared with the vanilla Faster R-CNN detector that it is built upon. In other words, switching to WTN from \textit{conventional classification weights} decreases performance on the seen classes. This phenomenon has been observed by prior works \cite{hoffman2014lsda,kang2018few} attempting to scale object detection with weak or partial supervision. By integrating autoencoder into WTN (AE-WTN), the seen-class detection performance can be recovered. It is extremely challenging to train conventional WTN from scratch. The reconstruction loss (which is more easily optimized than detection loss) encourages AE-WTN to output weights highly representative of original \clsnet weights, thus providing a good initialization to attain better local optima. Similar to prior works that find reduced supervised training loss with autoencoder \cite{le2018supervised,zhang2016augmenting}, we find that the box-level classification training loss \losscls \phantom{a}on seen classes attained by AE-WTN (0.5572) is lower than that of \wtnplus (0.5754).

Moreover, the reconstruction loss explicitly involves novel classes \novelcls during training and forces AE-WTN to preserve rich class information of the novel classes in the latent and output spaces. It also encourages visual features learned by \detnet to be more ``generic'' (less specific to classes \detnetcls in the detection dataset) in order to accommodate to the many classes represented by AE-WTN. Therefore, the detector equipped with AE-WTN shows improved (absolute) performances of $1.8\%$ and $1.1\%$ over \wtnplus on OI-57 and VG-200 respectively. Compared to other existing regularization techniques applied to \wtnplus, AE-WTN performs better across all datasets. This provides confirmation that the advantages of the reconstruction loss cannot be simply replicated by other regularizers that do not leverage the rich class information contained in \clsnet's weights.

\begin{figure*}[h]
\centering
\setlength{\tabcolsep}{1pt}
\begin{tabular}{ccc}
 \includegraphics[height=167px]{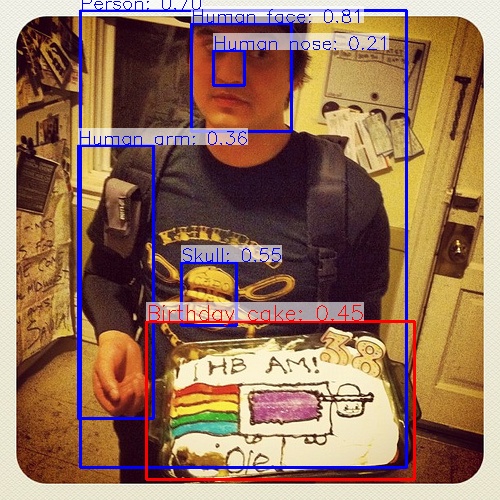}
\includegraphics[height=167px]{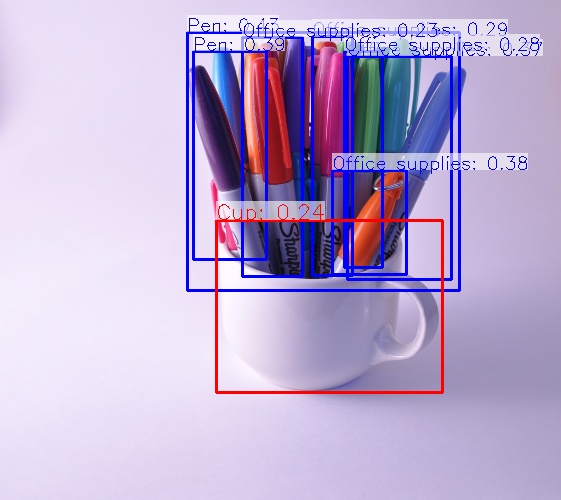}
\includegraphics[height=167px]{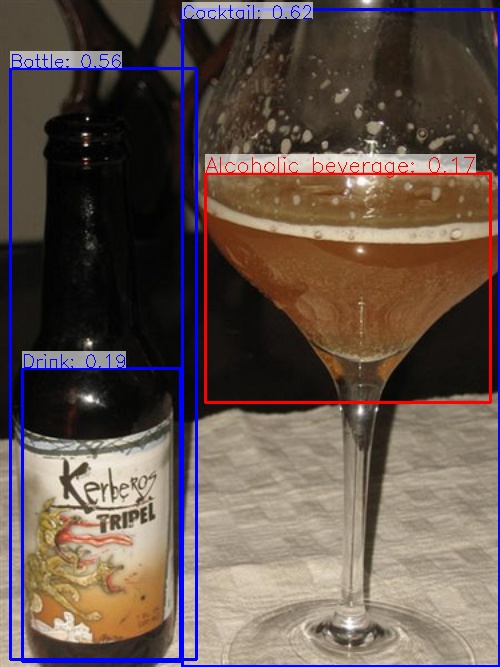}
\end{tabular}
\begin{tabular}{cc}
 \includegraphics[height=162px]{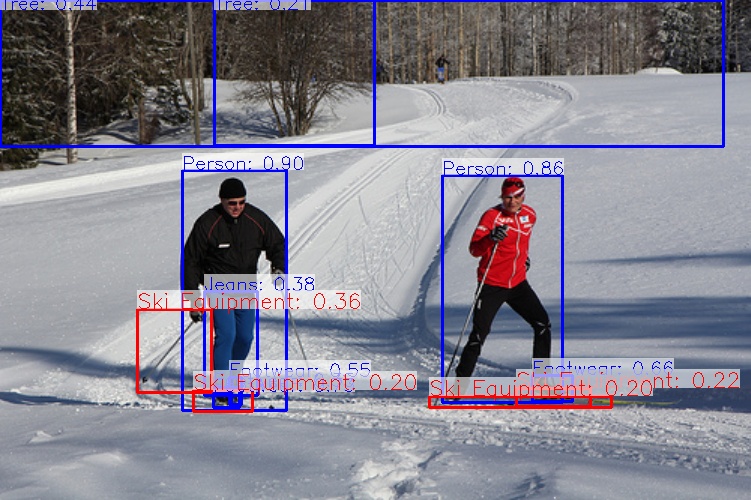}
\includegraphics[height=162px]{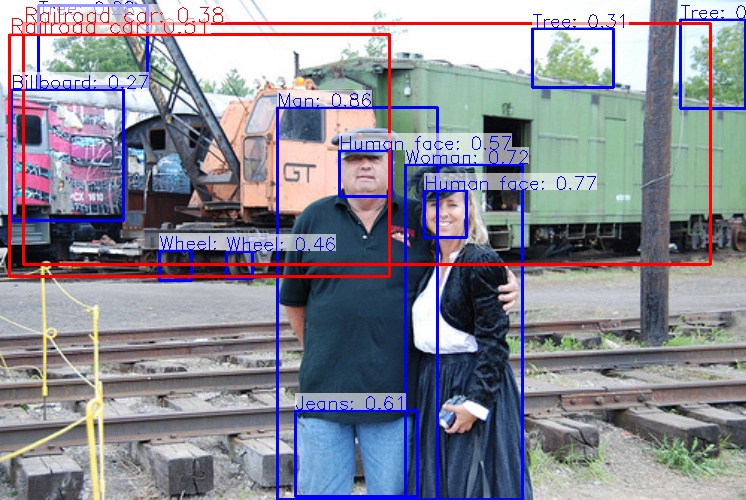}
\end{tabular}
\begin{tabular}{cc}
 \includegraphics[height=161px]{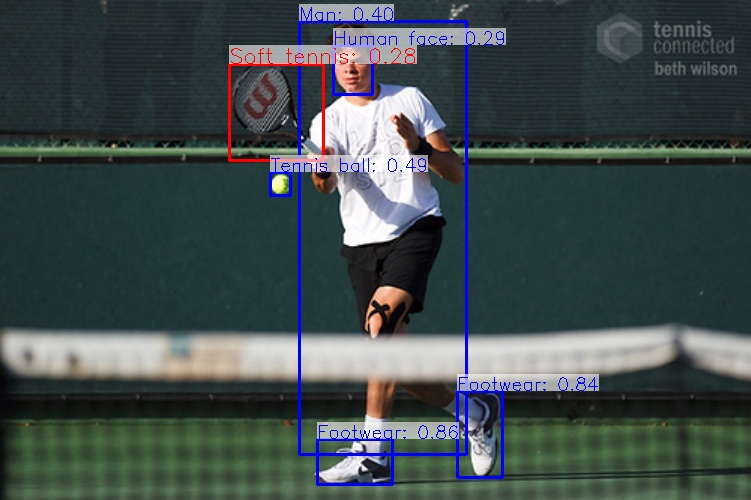}
\includegraphics[height=161px]{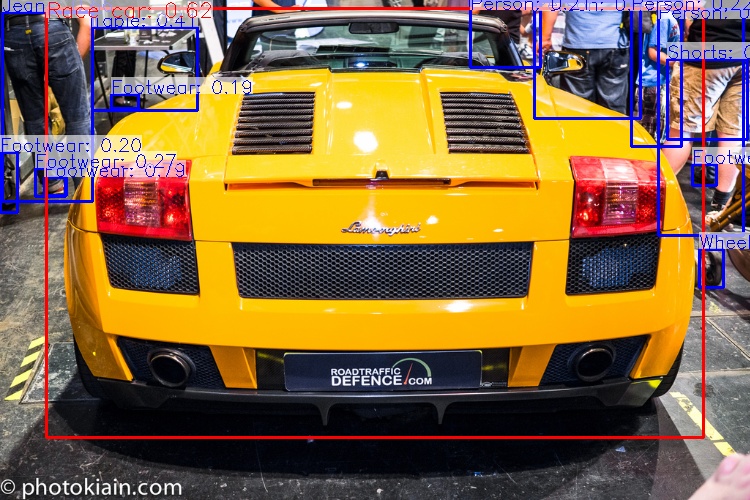}
\end{tabular}
	\caption{Qualitative results of the proposed AE-WTN. The \textcolor{blue}{blue} detection boxes \textbf{seen} object classes \detnetcls, and the \textcolor{red}{red} boxes labels are \textbf{novel} classes \novelcls. Our method can handle a variety of novel classes and concepts, while performing well on seen classes.}
	\label{fig:qualitative}
\end{figure*}

\noindent\textbf{Qualitative results}. We provide in Fig.~\ref{fig:qualitative} some qualitative results obtained by our proposed AE-WTN detector on test images of Open Images \cite{openimages} and Visual Genome \cite{krishna2017visual} datasets. Only the classes with the highest scores are shown, and novel classes compete with seen classes for the same bounding box. Remarkably, the detector can detect a variety of novel classes at greater confidence than seen classes, despite not seeing them during training.

\subsection{Analysis}

\begin{figure}[t]\centering
\includegraphics[width=\linewidth]{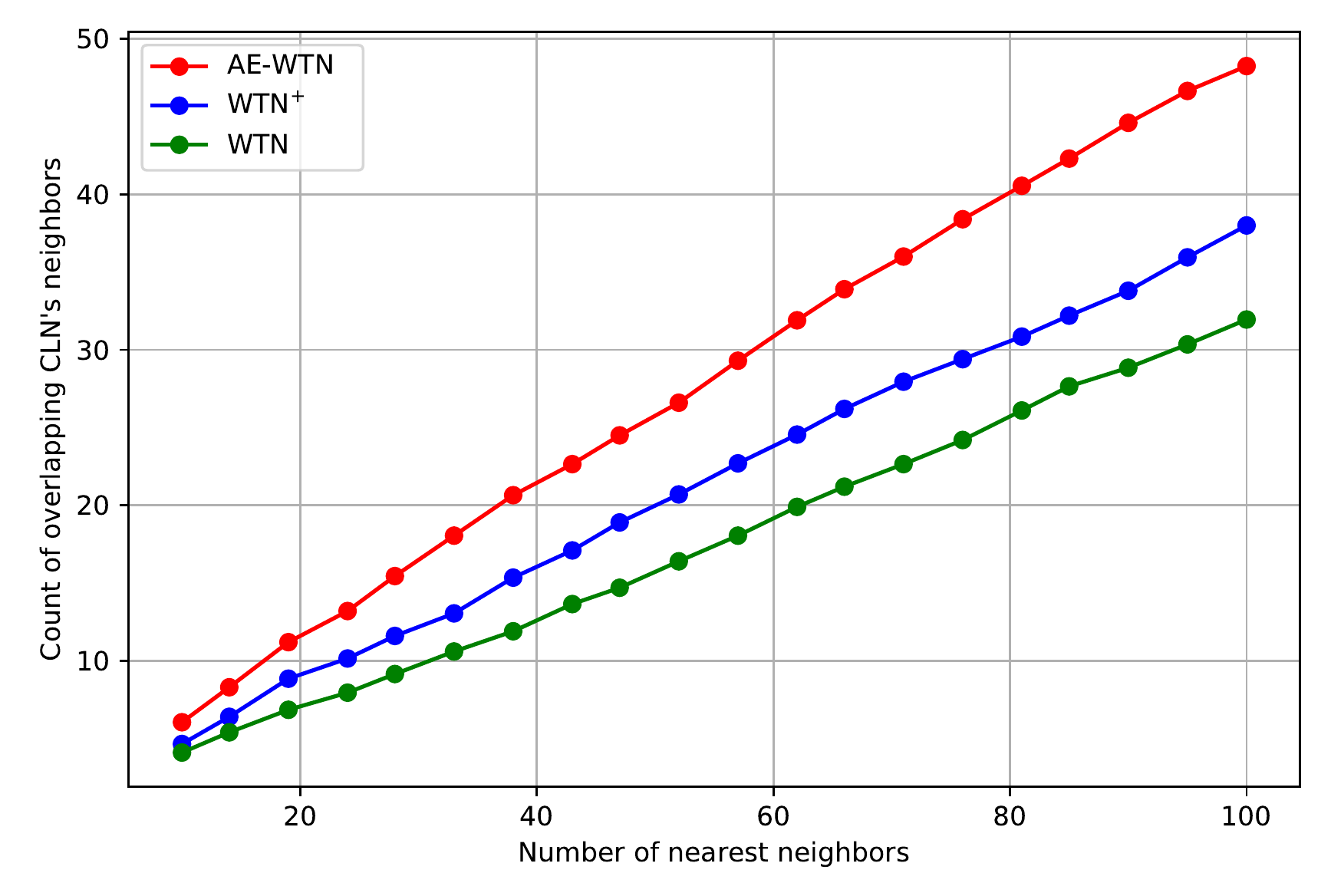}
	\caption{The overlapping count (vertical axis) between CLN's nearest neighbors and the nearest neighbors obtained by the WTN model of interest (AE-WTN, \wtnplus, or WTN), given varying numbers of nearest neighbors (horizontal axis).}
	\label{fig:overlapcount}
\end{figure}

\noindent\textbf{Local neighborhood preservation.} To better understand the implications of the reconstruction loss on local neighborhood preservation of AE-WTN, we compute the overlapping count between nearest neighbors obtained by \clsnet's weights and the output weights of the WTN model of interest (AE-WTN, \wtnplus, or WTN), varying the number of neighbors (a standard hyperparameter of \textit{nearest neighbor} approach) for all methods. This study is performed on 20 randomly-sampled classes and the counts are averaged across those classes. Nearest neighbors are among the 5,000 classes of \clsnet. The findings are presented in Fig.~\ref{fig:overlapcount}. E.g., at 100 neighbours, AE-WTN's output weights and \clsnet's weights have an average of 48.25 overlapping neighbours, while \wtnplus and WTN have 38.0 and 31.95 overlapping neighbours respectively. As shown, AE-WTN consistently reaches greater numbers of overlapping neighbors (with \clsnet's neighbors) than \wtnplus and WTN do, indicating that AE-WTN can better preserve the local neighborhood relationships of classes than \wtnplus. Noticeably, the gap widens as the number of nearest neighbors increases.\\

\noindent\textbf{Normalizations in \wtnplus.} We perform ablation study in Table \ref{table:ablation} to understand how the performance changes with different normalization techniques. It is crucial to combine the two normalizations of \wtnplus to obtain the best results for both seen and novel classes. Furthermore, we observe worse training losses with non-normalized WTN compared with \wtnplus, implying that model under-fitting is the inherent cause of WTN's under-performance.\\

\begin{table}[t!]
    \centering
    \begin{tabular}{c|c||c|c|c}
     \multicolumn{2}{c||}{} & \textbf{OI-500} & \textbf{OI-57} & \textbf{VG-200}\\
     \multicolumn{2}{c||}{\textbf{WTN}$\rightarrow$\textbf{\wtnplus}} & (Seen) & (Novel) & (Novel)\\\hline
     \begin{tabular}[c]{@{}l@{}}\textbf{Input}\\\textbf{Norm.}\end{tabular} & \begin{tabular}[c]{@{}l@{}}\textbf{Group}\\\textbf{Norm.}\end{tabular} & \ap                    & \ap                   & \ar                  \\\hline
    \xmark & \xmark & 52.87 & 34.94 & 41.91\\
    \cmark & \xmark & 57.60 & 37.27 & 54.19\\
    \xmark & \cmark & 54.60 & 35.84 & 58.55\\
    \cmark & \cmark & \textbf{58.82} & \textbf{39.28} & \textbf{65.60}\\
    \end{tabular}
    \vspace{5pt}
    \caption{Ablation study on \wtnplus architecture.}
    \label{table:ablation}
\end{table}

\noindent\textbf{Choice of feature normalization.} GN \cite{wu2018group} is chosen over the typical BatchNorm (BN) \cite{ioffe2015batch} because for \wtnplus, BN is less robust towards novel-class inputs which do not have detection annotations/loss \cite{galloway2019batch}. We find that the post-ReLU activation (L2) norms of \wtnplus with BN have an unusually large \textit{variance} for novel classes. It is $70$$\times$ (or $\frac{7.117}{0.104}$) that of shared classes, despite allowing BN to normalize over all classes in training. Such unstable activations are not encountered by the detection network during training. This causes \wtnplus's predicted weights for novel classes to interact poorly with image-region features at test time, resulting in unreliable class-score predictions. Table \ref{table:gnbn} shows  the L2 norm \textit{means} \& \textit{variances} of using GN and BN.\\
\begin{table}[t!]
    \centering
   \resizebox{\linewidth}{!}{
\begin{tabular}{c||cc|cc}
        & \multicolumn{2}{c|}{\textbf{mean}} & \multicolumn{2}{c}{\textbf{variance}}  \\\cline{2-5}
        & \textbf{shared cls.} & \textbf{novel cls.}           & \textbf{shared cls.} & \textbf{novel cls.}                \\\hline
GN \cite{wu2018group} & 1.838 & 1.784 & 0.091 & 0.093\\
BN \cite{ioffe2015batch} & 1.379 & \textbf{2.627} & 0.104 & \textbf{7.117}\\
\end{tabular}
    }
    \vspace{3pt}
    \caption{\textit{Means} and \textit{variances} of post-ReLU activation norms.}
\label{table:gnbn}
\end{table}

\begin{table}[t!]
    \centering
    \resizebox{\linewidth}{!}{
    \begin{tabular}{c||c|c|c|c}
    & \textbf{Faster R-CNN} & \textbf{WTN} & \textbf{\wtnplus} &
    \textbf{AE-WTN}\\\hline
    Time (ms) & 365 & 371 & 379 & 401\\
    Mem. (GB) & 4.11 & 4.15 & 4.19 & 4.26\\
    \end{tabular}
    }
    \vspace{3pt}
    \caption{Training time and memory usage.}
    \label{table:timemem}
\end{table}

\noindent\textbf{Computational efficiency.} Computational efficiency is a major concern in the training and/or deployment of object detectors, especially for large-scale detectors. In Table \ref{table:timemem}, we show the per-iteration training time (in milliseconds/\textit{ms}) and single-GPU memory usage of training with different models. Overall, the WTN models add very little computational costs on top of Faster R-CNN's. During testing, all the weights can be transformed offline with WTN/\wtnplus/AE-WTN to reach vanilla Faster R-CNN's efficiency.

%% file: tex/conclusion.tex
\section{Conclusion}
Training large-scale object detectors is extremely resource-demanding (e.g., data, computations). In this work, we introduce an efficient and effective WTN approach to scale up object detection, and propose novel methods to strongly push the limits of WTN through normalization techniques and autoencoder-based reconstruction loss. The reconstruction loss adopted by AE-WTN effectively improves its capability to retain and exploit the semantically-rich class information (of all classes) learned by the pre-trained \clsnet. This leads to improved training of \detnet and better detection performances on both seen and novel classes.

\section*{Acknowledgement}
\noindent Jason Kuen was supported by NTU Research Scholarship under School of Electrical and Electronic Engineering, Nanyang Technological University, Singapore.